\documentclass[letterpaper, 10 pt, conference]{ieeeconf}  

\IEEEoverridecommandlockouts                              

\usepackage{booktabs}

\usepackage[noadjust]{cite}
\nocite{*}  

\makeatletter
\let\NAT@parse\undefined
\makeatother

\usepackage{hyperref}
\usepackage{balance}
\usepackage{comment}
\hypersetup{
    colorlinks=true,
    citecolor=Green,
    linkcolor=NavyBlue,
    filecolor=OrangeRed,      
    urlcolor=OrangeRed,
    pdfpagemode=FullScreen,
}

\newcommand{\includegraphicsdpi}[3]{
    \pdfimageresolution=#1  
    \includegraphics[#2]{#3}
    \pdfimageresolution=20  
}
\usepackage{balance}
\usepackage{graphicx}
\usepackage{hyperref}
\usepackage{keyval}
\usepackage{algorithm}
\usepackage{algpseudocode}
\usepackage{amsmath, amsfonts}
\usepackage{amssymb}
\usepackage{cleveref}
\usepackage{fancyhdr} 
\usepackage{color}
\usepackage{transparent}
\usepackage{booktabs}
\usepackage{tabularx}

\usepackage{caption}
\usepackage{subcaption}
\usepackage{textcomp}
\usepackage{stfloats}
\usepackage{url}
\usepackage{verbatim}
\usepackage{graphicx}
\usepackage[table, dvipsnames]{xcolor} 
\usepackage{multirow}
\usepackage{mathtools}
\let\labelindent\relax
\usepackage{enumitem} 
\usepackage{booktabs}

\usepackage{caption}
\usepackage{tabularx}
\usepackage{threeparttable}
\usepackage{makecell}

\newif\ifshowrevs
\showrevstrue

\newif\ifshowauthorcomment
\showauthorcommenttrue

\newif\ifkeep
\keepfalse

\newif\ifremvspace
\remvspacetrue

\newcommand{\idest}{i.e., }
\newcommand{\exempli}{e.g., }

\newcommand{\etc}{etc.}
\newcommand{\safeshift}{\texttt{SafeShift}}

\newif\ifshowstatshorizontal
\showstatshorizontaltrue

\newif\ifshowauthorcomment
\showauthorcommenttrue

\newcommand{\cut}[1]{}

\ifshowrevs

\else

\fi

\usepackage{xargs} 
\usepackage[colorinlistoftodos,prependcaption,textsize=tiny]{todonotes}
\newcommandx{\unsure}[2][1=]{\todo[linecolor=red,backgroundcolor=red!25,bordercolor=red,#1]{#2}}
\newcommandx{\change}[2][1=]{\todo[linecolor=blue,backgroundcolor=blue!25,bordercolor=blue,#1]{#2}}
\newcommandx{\info}[2][1=]{\todo[linecolor=OliveGreen,backgroundcolor=OliveGreen!25,bordercolor=OliveGreen,#1]{#2}}
\newcommandx{\improvement}[2][1=]{\todo[linecolor=Plum,backgroundcolor=Plum!25,bordercolor=Plum,#1]{#2}}
\newcommandx{\thiswillnotshow}[2][1=]{\todo[disable,#1]{#2}}
 
\def\HiLiYellow{\leavevmode\rlap{\hbox to \hsize{\color{yellow!20}\leaders\hrule height .8\baselineskip depth .4ex\hfill}}}
  
\fancypagestyle{firststyle}
{
  \fancyhf{}

}

\fancypagestyle{secondstyle}
{
  \fancyhf{}
  \fancyhead[R]{\footnotesize \thepage}

}

\usepackage{cleveref}

\graphicspath{{figures/}}

\title{\LARGE \bf
SafeShift: Safety-Informed Distribution Shift for Robust \\Trajectory Prediction in Autonomous Driving
}

\author{Benjamin Stoler$^{1*}$ \and Ingrid Navarro$^{1*}$ \and Meghdeep Jana$^{1}$ \and Soonmin Hwang$^{2}$ \and Jonathan Francis$^{3}$ \and Jean Oh$^{1}$
\thanks{$^{*}$First authors BS and IN contributed equally. Work done by MJ and SH while at CMU. $^{1}$School of Computer Science, Carnegie Mellon University, Email: {\tt\footnotesize \{benstoler, jeanoh\}@cmu.edu, ingridn@cs.cmu.edu}.
$^{2}$Department of Automotive Engineering, Hanyang University, Email: {\tt\footnotesize soonminh@hanyang.ac.kr}.
$^{3}$Bosch Center for Artificial Intelligence, Email: {\tt\footnotesize jon.francis@us.bosch.com}.}
}

\let\oldtwocolumn\twocolumn
\renewcommand\twocolumn[1][]{%
    \oldtwocolumn[{#1}{
    \begin{center}
           \includegraphicsdpi{50}{width=1.0\textwidth}{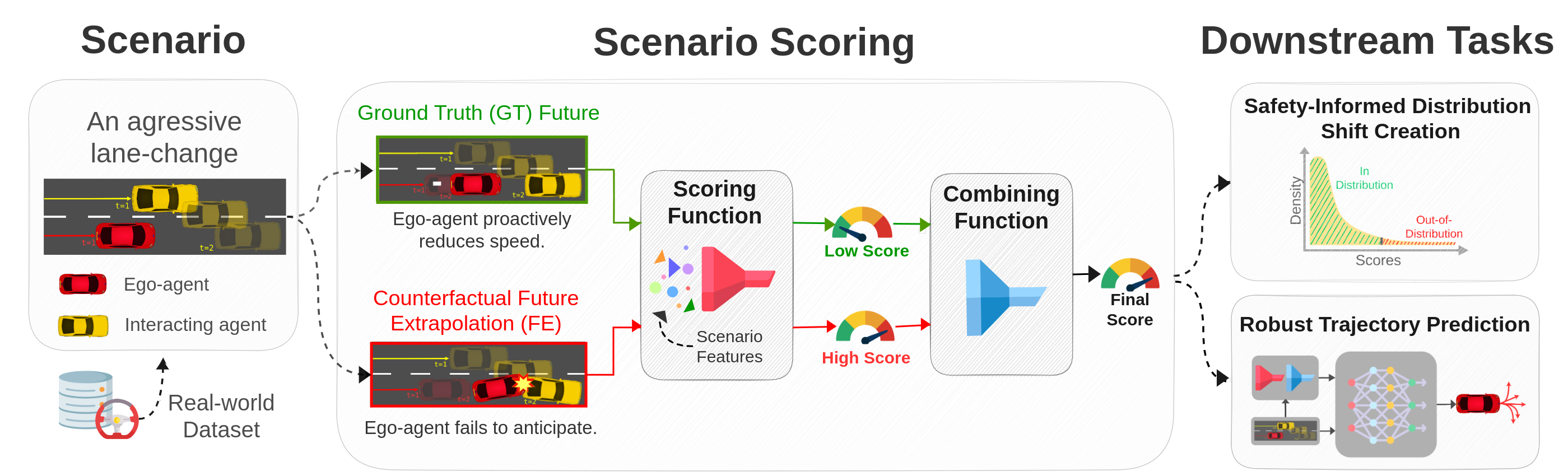}
           \vspace{-0.2cm}
           \captionof{figure}{An overview of \safeshift. Our framework consists of a scoring methodology that uses counterfactual probing to characterize and score scenarios, exploring \textit{what-if} scenarios where proactive maneuvers were not performed, thus resulting in safety-criticality or near misses. We also apply and assesses this scoring approach on two downstream tasks: safety-informed \textit{distribution shift creation}, where challenging scenarios are found and held out for evaluation; and \textit{robust trajectory prediction}, where trajectory prediction algorithms are assessed under this distribution shift and remediated.}
           \label{fig:overview}
        \end{center}
    }]
}

\begin{document}

\maketitle
\thispagestyle{empty}
\pagestyle{empty}

\begin{abstract}

As autonomous driving technology matures, the safety and robustness of its key components, including trajectory prediction is vital. Although real-world datasets such as Waymo Open Motion provide recorded real scenarios, the majority of the scenes appear benign, often lacking diverse safety-critical situations that are essential for developing robust models against nuanced risks. However, generating safety-critical data using simulation faces severe simulation to real gap. Using real-world environments is even less desirable due to safety risks. In this context, we propose an approach to utilize existing real-world datasets by identifying safety-relevant scenarios naively overlooked, e.g., near misses and proactive maneuvers. Our approach expands the spectrum of safety-relevance, allowing us to study trajectory prediction models under a safety-informed, distribution shift setting. We contribute a versatile scenario characterization method, a novel scoring scheme for reevaluating a scene using counterfactual scenarios to find hidden risky scenarios, and an evaluation of trajectory prediction models in this setting. We further contribute a remediation strategy, achieving a 10\% average reduction in predicted trajectories' collision rates. To facilitate future research, we release our code for this overall SafeShift framework to the public: \url{github.com/cmubig/SafeShift}

\end{abstract}


\section{Introduction} \label{sec:introduction}


As autonomous driving (AD) technologies are increasingly deployed in the wild, the safety and robustness of the autonomous systems remain chief concerns~\cite{guo2022maturity, francis2022learn, teng2023motion}. One key AD task is that of trajectory prediction, wherein the future trajectories of agents in a scene must be predicted, given a brief historical observation. These predictions may be used in the downstream portion of conventional vehicle control stacks, to inform an ego-agent's motion planner as it attempts to find possible conflict-free and traffic infraction-free paths. Thus, improving the agent's robustness and its ability to detect possibly safety-critical scenarios is of paramount importance in ensuring the overall acceptable performance of autonomous vehicles in real-world deployments~\cite{kothari2023safety}.

It is appealing to train trajectory prediction models using large real-world motion prediction datasets, such as the 
Waymo Open Motion Dataset (WOMD)~\cite{ettinger2021large}, 
as they consist of recorded scenarios capturing the behaviors of various agents---human drivers and vulnerable road users (VRUs), e.g., cyclists and pedestrians---under real-world traffic layouts and densities. 
One inherent challenge in using such datasets, however, is that the frequency of vehicle infractions and other safety-critical scenarios therein is quite low. 
The prior art regards this issue as the ``curse of rarity''~\cite{ding2020learning, ding2023survey, feng2023dense} and, as a result, industry and academia have resorted to validating autonomous driving agents via on-road tests~\cite{webb2020waymo, huang2016autonomous}, where those valuable rare events are also potentially dangerous to other drivers and VRUs, or via simulated experiments~\cite{ding2020learning, ding2023survey, feng2023dense}, wherein the artificial behaviors of agents and inaccurate world physics in the simulators can leave models unprepared and inadequate for real-world deployment~\cite{francis2022core, huang2023went}. 

Recently, several works have identified a potential solution to this challenge of robust training, by generating ``new'' traffic scenes that serve as training samples for otherwise rare events and/or as difficult test-cases to challenge already-trained models. 
Unfortunately, despite recent advances in safety-critical scenario generation methods~\cite{xu2023bits, cao2023robust, suo2023mixsim}, generating non-trivially challenging cases that match the realism, frequency, and difficulty of safety-critical scenarios that agents might encounter in the real world remains an open problem. 
An effective and under-explored alternative lies somewhere in the middle: 
we propose an approach to mine large-scale datasets of real-world vehicle deployments to find and leverage meaningful \textit{safety-relevant} scenarios that may be hidden in the data. Our key insight is that, in autonomous driving, safety-relevance includes not just scenarios where observed agents act in a safety-critical\cut{or near safety-critical} manner, but also scenarios where agents are able to avoid infractions through\cut{more subtle and} proactive maneuvers. 
Therefore, we propose to leverage counterfactual probes to additionally characterize \textit{what-if} scenarios where these proactive maneuvers were \textit{not} performed. Such fine-grained scenario characterization enables trajectory forecasting models to more easily distill diverse defensive driving skills \cite{shalev2017formal} from existing datasets, e.g., preemptive braking as illustrated in \Cref{fig:overview} (left).



Under this paradigm of scenario characterization, we propose the \safeshift~framework for identifying and studying the most safety-relevant scenarios in a widely-available autonomous driving dataset. 
The more extreme scenarios are held out as an out-of-distribution (OOD) test-set, thus acting as a stand-in for the valuable and rare, long-tailed events. In this way, we avoid both the challenges of attempting to generate new safety-critical scenarios as well as the challenges in performing simulation-to-real transfer; instead, we optimize the usefulness of existing data. To the best of our knowledge, prior work that focuses on creating artificial distribution shifts have not done so based on safety-relevance, instead focusing on, \exempli lane or global location characteristics~\cite{filos2020can, ye2023improving}, speed of driving~\cite{itkina2023interpretable}, or the city that the data was captured in~\cite{itkina2023interpretable}. Furthermore, prior efforts in scenario characterization under distribution shift settings rely on empirical, dataset-specific heuristics~\cite{moers2022exid, glasmacher2022automated, sadat2021diverse}. 

Our main contributions, illustrated in \Cref{fig:overview}, are thus as follows: 
1) A versatile approach for scenario characterization in autonomous driving, focused on capturing safety-relevant scenarios; 
2) A methodology for scoring safety-criticality for the purposes of crafting a distribution shift, including novel progress in incorporating the aforementioned fuller spectrum of safety-relevance, and improving safety performance therein; and 
3) An evaluation of existing socially-aware trajectory prediction approaches in this safety-informed distribution shift setting, utilizing WOMD~\cite{ettinger2021large} as an exemplar. Our developed remediation strategy for this setting reduces the predicted trajectories' collision rates by an average of $10\%$, across the tested models. 

\section{Related Work} \label{sec:related_works}

\subsection{Socially-Aware Trajectory Prediction}

Motion prediction in crowded environments is a well-researched task in the domains of autonomous driving and motion in human crowds~\cite{rudenko2020human}. Most current approaches for motion prediction are data-driven, \idest they focus on characterizing behavior and interactions observed in the data.
To capture a multi-modal distribution of possible futures, generative frameworks are frequently used~\cite{monti2020dagnet, sadeghian2019sophie, park2020diverse, shi2022motion, bhat2020trajformer, tang2022golfer}. To model joint behavior and social cues, various techniques such as social pooling~\cite{alahi2016sociallstm}, rasterized representations~\cite{konev2022motioncnn}, and attention-based methods~\cite{monti2020dagnet, yan2021agentformer, ngiam2021scene, shi2022motion} have been employed. Several state-of-the-art techniques have also explored learning richer representations for motion prediction, such as modeling context information as road graphs or polylines~\cite{liang2020learning, gao2020vectornet} and goal conditioning~\cite{sadeghian2019sophie}.


\subsection{Robustness Assessment in Trajectory Prediction}

One approach to examine robustness for trajectory prediction is robustness to adversarial attacks. Recent studies have shown that state-of-the-art prediction models often lack basic social awareness and collision avoidance when faced with these attacks~\cite{weng2023joint, saadatnejad2022socially, zhang2022adversarial}. A significant disadvantage with these techniques however is that they ultimately rely on simulating realistic agent behavior, which often incurs a simulation-to-real gap~\cite{hanselmann2022king, xu2023bits, francis2022core, huang2023went}. Another approach to ensuring robustness involves studying models' performances under a data domain distribution shift setting, recognizing that AD models will ultimately encounter unseen scenarios in the wild.
Some approaches involve identifying domains based on meta characteristics of the scene, such as road shape characteristics, side-of-driving, and average speeds~\cite{filos2020can, itkina2023interpretable}. Another recent method explores clustering scenes into domains based on several features, including lane deflection angles, global bounds of the scenario and trajectories, and lane shape information~\cite{ye2023improving}. Many of these works also include domain adaptation or remediation strategies to reduce the impact of the distribution shift, such as by leveraging Frenet coordinates~\cite{ye2023improving, werling2010optimal}, few-shot adaptation~\cite{filos2020can}, or motion-based style transfer~\cite{kothari2023motion}. However, to the best of our knowledge, no work has attempted to create distribution shifts based on safety-relevance or study remediation therein.

\subsection{Critical Scenario Identification in Autonomous Driving} 

Many existing datasets rely on mining the immense amount of collected data from road-tests for interesting scenes~\cite{ettinger2021large, chang2019argoverse, caesar2020nuscenes}, considering surface-level metrics such as traffic density and kinematic complexity. Therefore, prior frameworks for critical scenario identification (CSI) have been designed to expand upon these initial dataset characterizations~\cite{zhang2021finding, weber2019framework}. These frameworks typically focus on creating taxonomies for categorizing conflict scenarios, as well as for developing metrics and validation methods to describe and cope with them. 
In \cite{glasmacher2022automated}, scenarios are instead hierarchically scored along metrics related to anomaly, interestingness, and relevance, better handling more complex maneuvers. Another recent work expands beyond this by defining complexity aspects relating to the road graph layout, surrounding objects, and topology of agents' paths~\cite{sadat2021diverse}. However, the use of these surrogate metrics for CSI alone, without applying counterfactual reasoning, can fail to identify more subtle safety-relevant scenarios, as illustrated in \Cref{fig:overview}. Furthermore, these metrics often rely on empirical weighting and thresholding schemes, as well as on privileged information not uniformly available in AD datasets (\exempli global reference frames, drivable area identification, \etc)~\cite{glasmacher2022automated, sadat2021diverse}; thus they cannot be applied to several key datasets, including WOMD. 

\begin{figure*}[htbp]
    \vspace{0.2cm}
    \centering
    \hfill
    \begin{subfigure}[b]{0.45\textwidth}
        \includegraphics[width=\textwidth]{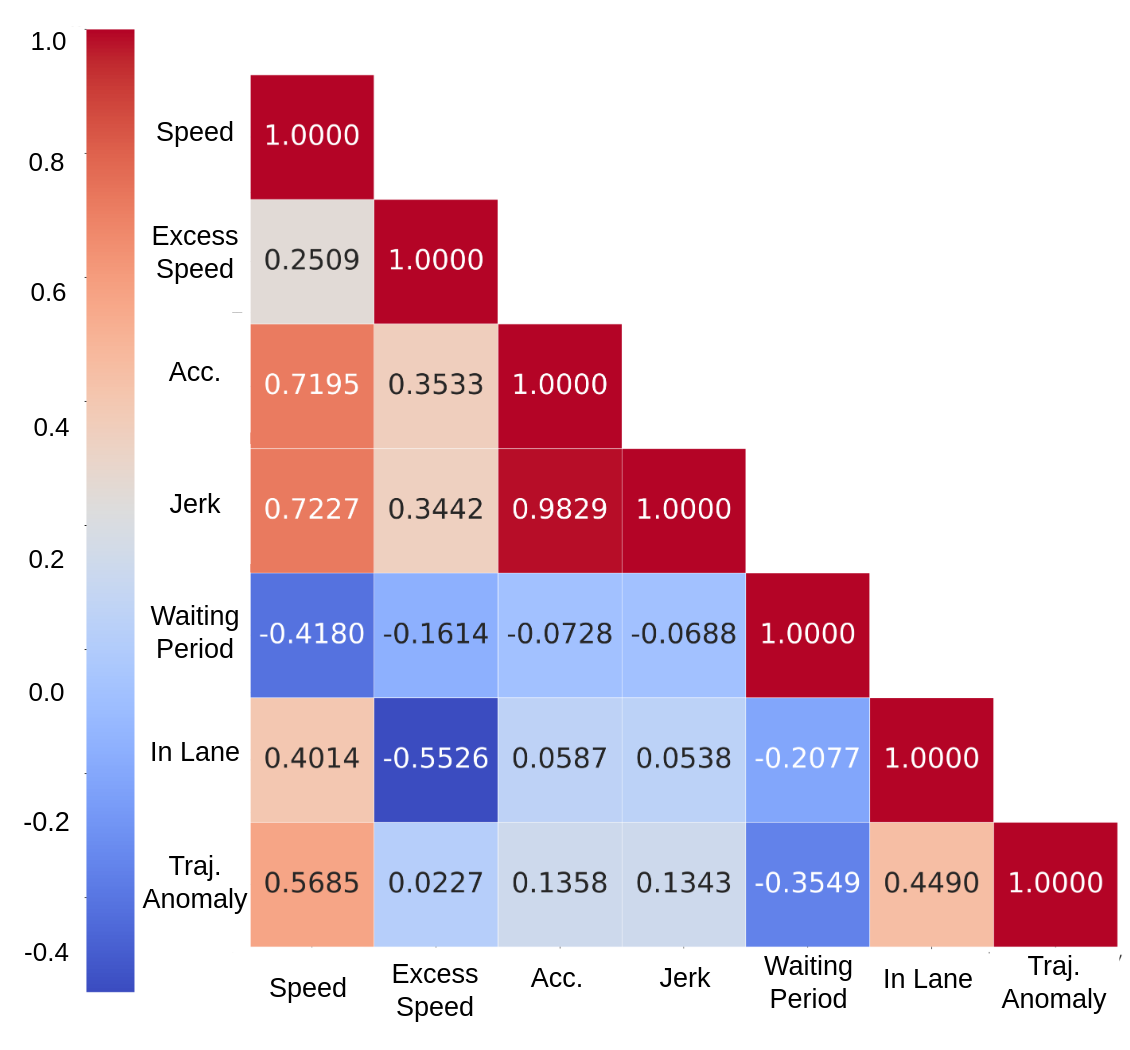}
        \caption{Correlation Analysis for Individual Features}
        \label{subfig:individual_corrmat}
    \end{subfigure}
    \hfill
    \begin{subfigure}[b]{0.45\textwidth}
        \includegraphics[width=\textwidth]{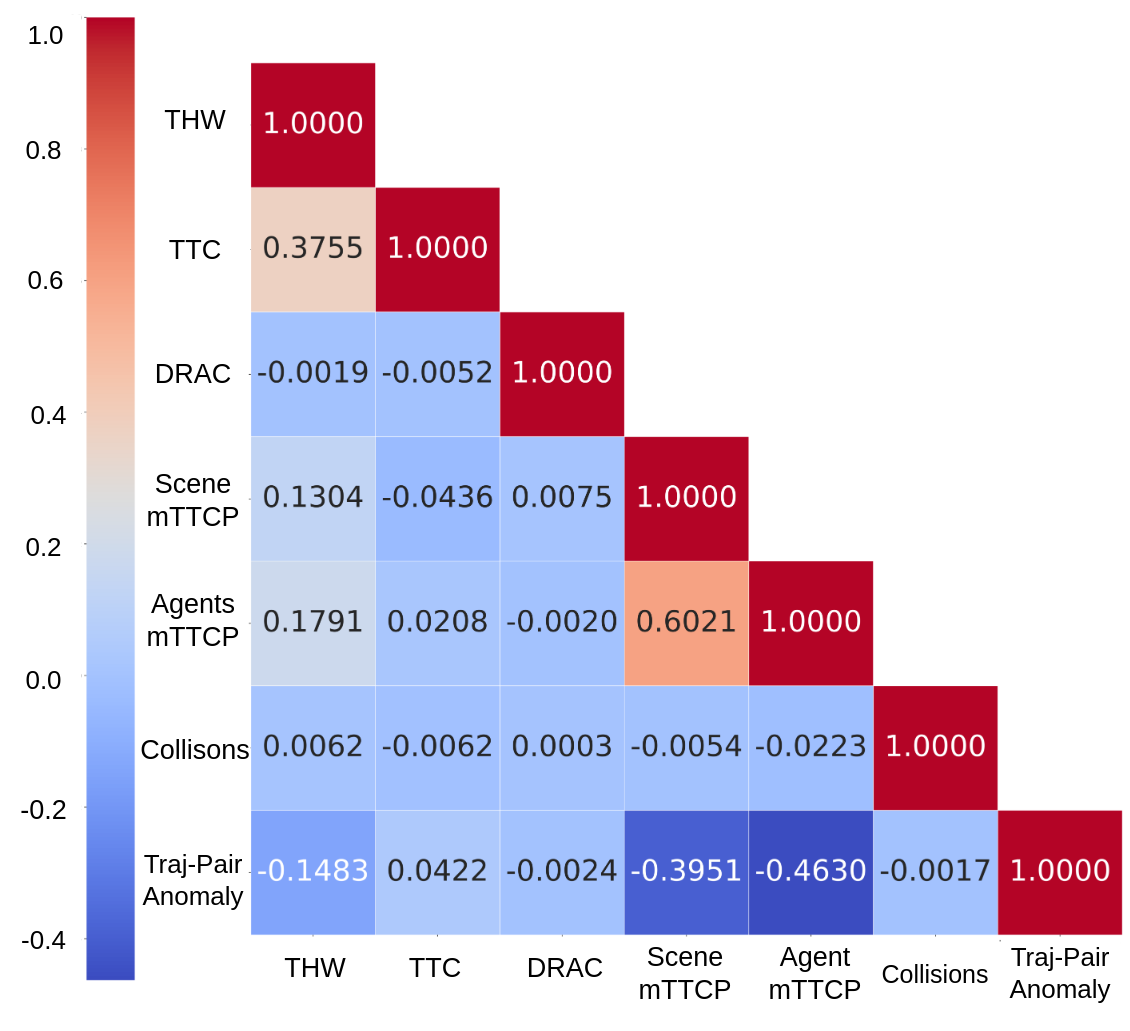}
        \caption{Correlation Analysis for Social Features}
        \label{subfig:interaction_corrmat}
    \end{subfigure}
    \hfill
    \caption{Pearson correlation coefficients for each pair of metrics, showing how the features complement each other. Analysis performed on WOMD~\cite{ettinger2021large}.}
    \vspace{-0.5cm}
    \label{fig:correlation_analysis}
\end{figure*}

\section{Preliminaries} \label{sec:problem_definition}

In this section, we define the task of trajectory prediction under distribution shifts. First, we consider the set of scenarios that comprise a motion prediction dataset as $\mathcal{S}$. Thus, we denote $s \in \mathcal{S}$ as a single scenario taken from this corpus. The scenario $s$ consists of all agent tracks $\mathbf{X}$, map information, and meta information. Agent tracks are the time-varying locations of every observed agent in the scene, in a Cartesian frame, where $\mathbf{X}^{(i)}_t$ denotes the location of agent $i$ at timestep $t$. The map contains road information, e.g., lane locations and lane connectivity, and the meta information provides additional task specifications, such as the list of which agents are to be predicted.\footnote{The exact format of $s$, such as the origin of the Cartesian frame or specific set of map information provided, varies from dataset to dataset.}

The information in $s$ is further split into a history and future portion:  $s_{hist} = \{s_1, s_2, ..., s_{T_{obs}}\}$ and $s_{fut} = \{s_{T_{obs + 1}}, s_{T_{obs + 2}}, ..., s_{T_{tot}}\}$, where $T_{obs}$ denotes the timestep before the prediction horizon and $T_{tot}$ denotes the total length of the scene. Thus, the task of trajectory prediction is to jointly estimate the values of $\mathbf{X}^{(i)}_{fut}$, using only $s_{hist}$, for all agents $i \in \{1,2,3,...\}$.

Under distribution shift conditions, $\mathcal{S}$ may be split into two sets---$\mathcal{S}_{ID}$, representing the in-distribution set, and $\mathcal{S}_{OOD}$ representing the out-of-distribution set. The task of robust trajectory prediction then, is to minimize the drop in performance on safety-relevant metrics for trajectory prediction (\idest collision rates) when prediction models are tested on $\mathcal{S}_{OOD}$, after being trained and validated only on $\mathcal{S}_{ID}$.

The remaining sections in this work are organized as follows. In \Cref{sec:approach} and \Cref{sec:score_scheme}, we propose a novel scenario-characterization framework and scoring methodology. Then in \Cref{sec:downstream_tasks} and \Cref{sec:experimental_setup}, we show how to leverage our framework for introducing safety-informed distribution shifts in a given autonomous driving dataset and for developing remediation strategies to improve the robustness of trajectory prediction models. Finally in \Cref{sec:results} we discuss the results and implications of these experiments.


\section{Scenario Features} \label{sec:approach}


We propose a hierarchical scheme as in~\cite{glasmacher2022automated, sadat2021diverse}, where low-level, base features are computed within a scenario and then later aggregated to form a score representing a scenario's overall safety-relevance. We consider base features across two main categories: \textit{individual} features related to single-agent behavior and \textit{social} features relevant to the interactions between agents. 


For both of these categories, accurate lane assignment is highly important but is nontrivial, e.g., VRUs often do not adhere to lanes.
Whereas a simple method of snapping to the best-fitting local lane has been used in previous work~\cite{ye2023improving}, we instead leverage a probabilistic approach~\cite{schmidt2022meat} to find valid lane \textit{sequences} for agents. Additionally, we permit lane assignments based on physically plausible lane deflection angles rather than the lane connectivity graph alone.
We excluded some features utilized in previous frameworks and datasets~\cite{glasmacher2022automated, moers2022exid}, such as driving region-based anomaly detection, that require the knowledge of global, city coordinates which are not generally available across all AD datasets. Instead, to identify anomalies, we utilize a traffic primitive extraction and clustering approach pioneered in \cite{guha2022robust}. This process produces cluster centers for both single trajectories and trajectory pairs, allowing us to easily measure anomalies.

\vspace{0.1cm}
\noindent\textbf{Individual Features}: We primarily focus on metrics derived from relative positional data of a trajectory, such as speed, acceleration, and jerk. We additionally implement metrics to incorporate map context, including waiting period (WP)~\cite{zhan2019interaction}, speed difference with the lane's speed limit, and the percentage of time that the agent is following a lane. Finally, we include a trajectory anomaly value, derived from its distance to the nearest individual traffic primitive cluster. 

\vspace{0.1cm}
\noindent\textbf{Social Features}: We use widely studied and accepted safety surrogate metrics, as in \cite{glasmacher2022automated, vogel2003comparison, shen2020crash}. These include time headway (THW), time-to-collision (TTC), deceleration rate to avoid crash (DRAC), and the difference between minimum time to conflict points ($\Delta$mTTCP) 
in both agent trajectories and road graph locations of interest (\exempli crosswalks, stop signs). 
We then incorporate a measure of collisions directly, counting situations where two agents' center points or segmented paths overlap at a given timestep. Finally, analogous to the individual trajectory anomaly, we add a trajectory-pair anomaly value using paired traffic primitive clusters.






\begin{figure}[t!]
    \vspace{0.2cm}
    \centering
    \includegraphics[width=0.49\textwidth,trim={0cm 0cm -0cm -0cm},clip]{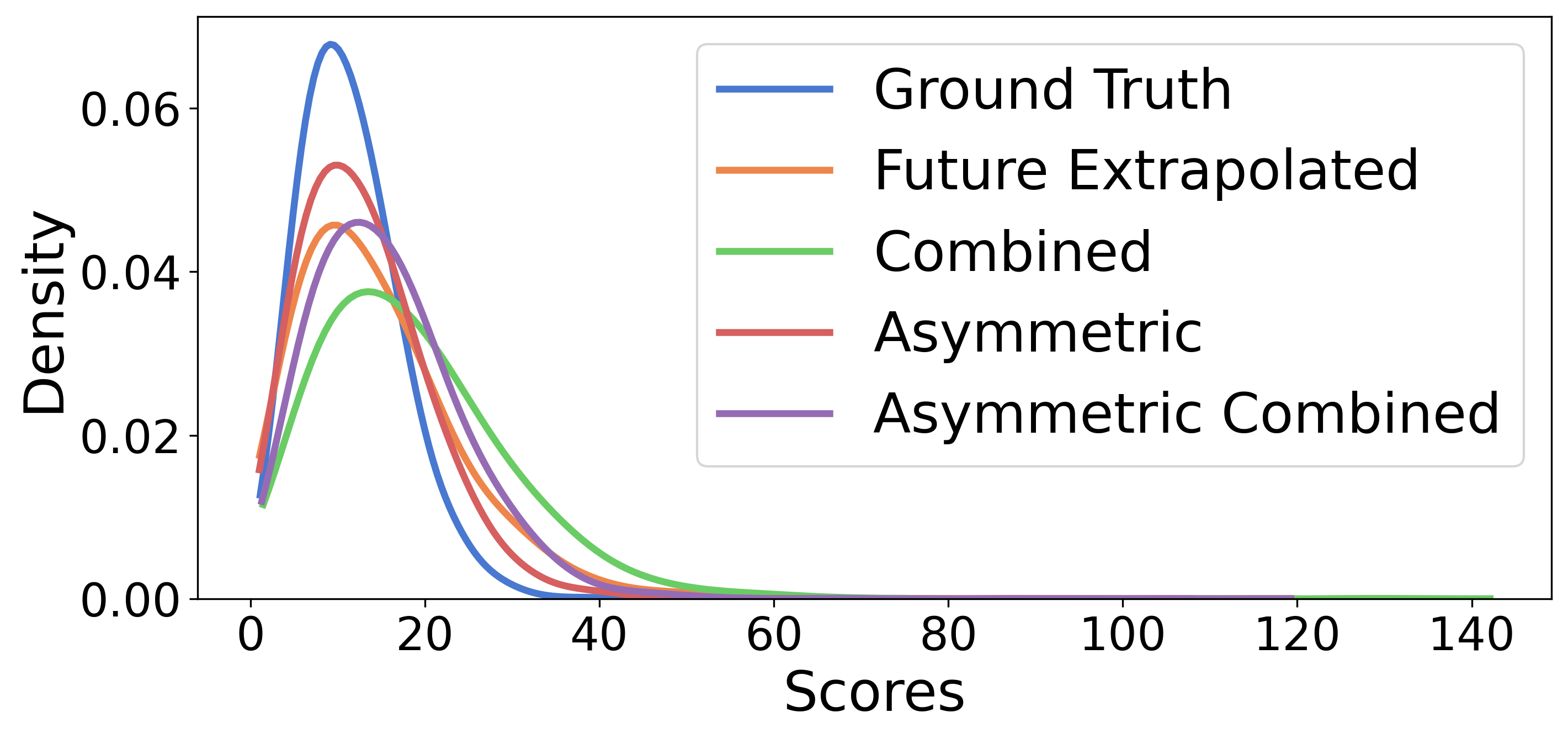}
    \vspace{-0.2cm}
    \caption{PDF of our score variations, exhibiting long-tailed behavior. Analysis performed in WOMD~\cite{ettinger2021large}.}
    \label{fig:score_dist}
\end{figure}

\begin{table}[t]
\vspace{0.1cm}
\centering
\caption{Trajectory scoring variations.} \label{tab:score_variations}
\resizebox{0.49\textwidth}{!}{%
\begin{tabular}{lcc}
\toprule

Variation & \texttt{IndScore} & \texttt{SocScore}  \\ 

\midrule
  
   \texttt{Ground Truth } $(GT)$  & $\mathbf{X}^{(i)}_{GT}$ & $(\mathbf{X}^{(i)}_{GT}$,  $\mathbf{X}^{(j)}_{GT})$  \vspace{0.1cm} \\
   
   \texttt{Future Extrapolated} $(FE)$ & $\mathbf{X}^{(i)}_{FE}$ & ($\mathbf{X}^{(i)}_{FE}$, $\mathbf{X}^{(j)}_{FE}$)  \vspace{0.1cm}\\
   
   \texttt{Asymmetric} $(AS)$ & $\mathbf{X}^{(i)}_{FE}$ & ($\mathbf{X}^{(i)}_{FE}$, $\mathbf{X}^{(j)}_{GT}$)  \vspace{0.1cm}\\
   
   \midrule
   
   \texttt{Combined} $(CO)$ & \multicolumn{2}{c}{
    $\max (\texttt{TrajScore}_{\texttt{GT}}, \texttt{TrajScore}_{\texttt{FE}}) $
   }   \vspace{0.1cm}\\
   
   \texttt{Asymmetric Combined} $(AC)$ & \multicolumn{2}{c}{
    $\max (\texttt{TrajScore}_{\texttt{GT}}, \texttt{TrajScore}_{\texttt{AS}}) $
   }   \vspace{0.1cm}\\
   
\bottomrule
\end{tabular}
}
\vspace{-0.2cm}
\end{table}


Our full feature selection, along with a correlation analysis is shown in \Cref{fig:correlation_analysis}. For the individual features, the kinematic-based ones correlate positively, as could be expected, while the other features are largely weakly correlated. Similarly, for the social features, TTC and THW have a weak correlation, as they both involve a leader-follower scenario. The two forms of $\Delta$mTTCP are also relatively strongly correlated, as agent trajectories are required to be somewhat intertwined for both. This analysis implies that the selection and extraction of base features are largely complementary, without excessive overlap in coverage. 

\begin{figure*}[t]
    \vspace{0.2cm}
    \centering
    \includegraphics[width=0.95\textwidth,trim={0cm 0cm -0cm -0cm},clip]{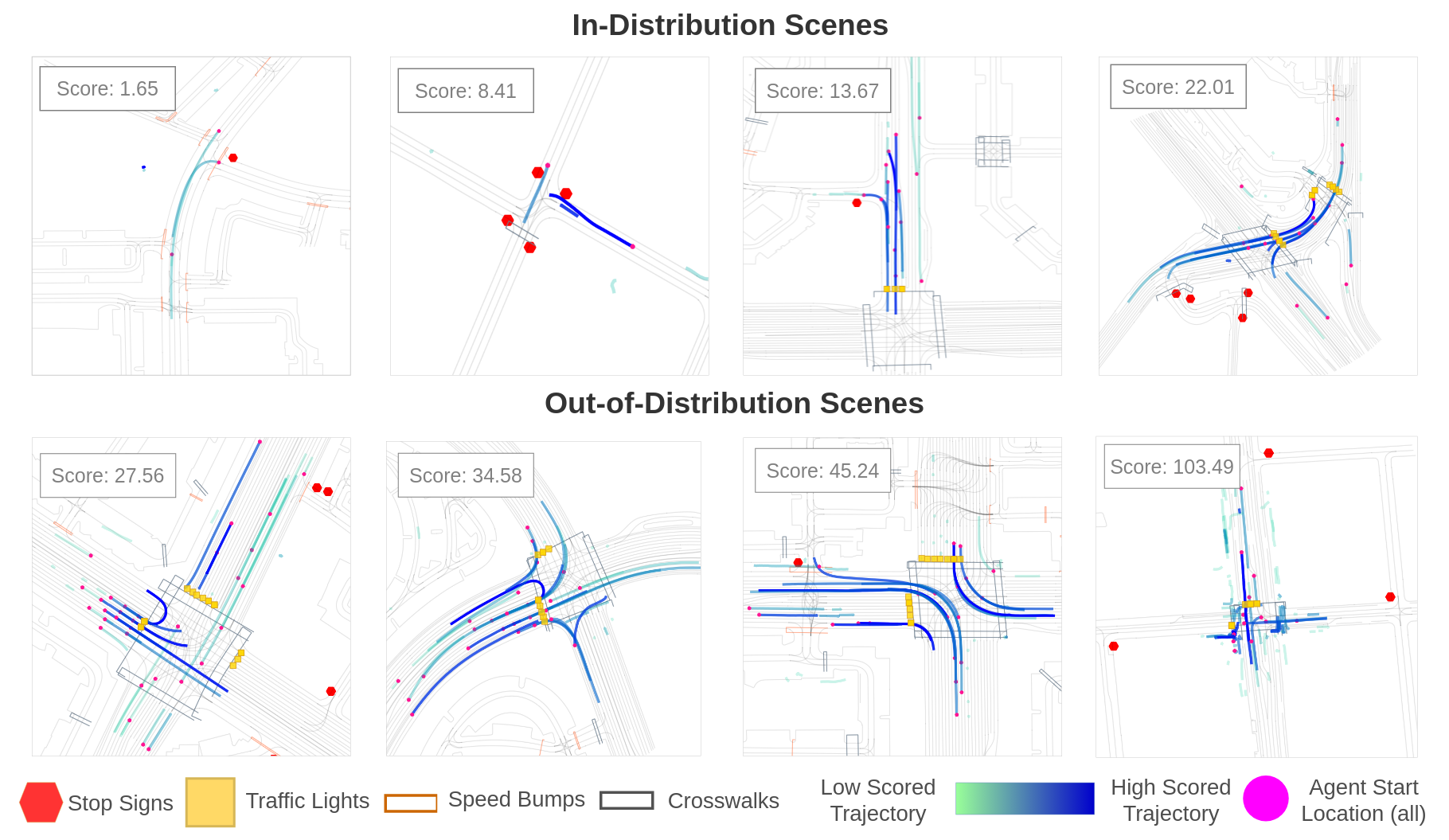}
    \caption{Examples of WOMD~\cite{shi2022motion} scenes by score. In-Distribution and Out-of-Distribution follow our \texttt{Scoring} split in \Cref{ssec:distribution_shift_creation}.}
    \vspace{-0.5cm}
    \label{fig:scored_scenes}
\end{figure*}

\section{Scenario Scoring}
\label{sec:score_scheme}

Using the base features described in \Cref{sec:approach}, we define a safety-relevance scoring function that can characterize a given scenario. We then propose a counterfactual re-scoring approach where we re-characterize the same scenario by taking \textit{what-if} alternatives into account. 


\subsection{Scoring Functions}

We start by hierarchically aggregating the base features to create overall trajectory and scene scores as follows. 
Let $\mathbf{V}_{ind}$ be the total set of extracted individual features, $\mathbf{V}_{soc}$ be the set of social features, and $v \in \mathbf{V}$ represent a single feature taken from one of these sets (\exempli acceleration, TTC, etc.). Then, let $v_{t}^{(i)}$ be such an extracted individual feature $v$ for trajectory $i$ at timestep $t$. Similarly, $v_{t}^{(i, j)}$ denotes a social feature over trajectories $i$ and $j$ together. 

To combine these extracted base features, we first convert them to a form in which a larger value corresponds to more safety-relevance (\idest for features such as speed, we use $v$ directly, but for features such as TTC, we use $1/v$). We then aggregate the individual features into an individual score. We take the maximum value for each metric incurred throughout the trajectory and then linearly combine them according to weights specified in \cite{glasmacher2022automated}; let these weights be denoted as $\mathbf{W}_{ind}$ and $\mathbf{W}_{soc}$. Then, a trajectory's individual score is expressed in \Cref{eq:individual_score}, where ``$\ \cdot \ $'' denotes the vector scalar product: %
\begin{equation}\label{eq:individual_score}
    \texttt{IndScore}^{(i)} = \mathbf{W}_{ind} \cdot \left[ \max_t (v_t^{(i)}) \ | \ v \in \mathbf{V}_{ind} \right]
\end{equation}

Note that we do not perform any sort of value detection thresholding to avoid reliance on empirical decision making. Similarly, for each pair of trajectories, we compute a social score, as follows in \Cref{eq:social_score}: %
\begin{equation}\label{eq:social_score}
    \texttt{SocScore}^{(i, j)} = \mathbf{W}_{soc} \cdot \left[ \max_t (v_t^{(i, j)}) \ | \ v \in \mathbf{V}_{soc}    
    \right]
\end{equation}

An agent's trajectory score is then computed by adding together its \textit{individual} score with the \textit{social} score of all trajectory pairs it is involved in: %
\begin{equation}\label{eq:traj_score}
    \texttt{TrajScore}^{(i)} = \texttt{IndScore}^{(i)} + \sum_{j \neq i} \texttt{SocScore}^{(i, j)}
\end{equation}

We combine these \texttt{TrajScores} into a final \texttt{SceneScore} as follows. We begin by taking the weighted sum of all agents' scores in the scene, where each weight is inversely proportionate to its minimum distance to an agent marked as requiring prediction. Then, to regularize the effect of scene density, we normalize this total, proportionate to the total number of agents present. 

\subsection{Counterfactual Re-Scoring} \label{ssec:counterfactual_rescoring}

The key insight of counterfactual re-scoring is to assess the safety-criticality of a scenario based on potential \textit{what-if} cases in addition to the recorded ground truth event. We hypothesize that the characterization using counterfactual scenarios can capture the hidden risks better than using the ground truth record only, which will subsequently result in improved performance in downstream tasks such as robust trajectory prediction.

To find scenarios beyond just those with high aggregated criticality and/or surrogate criticality values,
we wish to perform a counterfactual probe into what could happen if an agent were to simply maintain its current progress within a lane. This represents, \exempli the behavior of a distracted driver ignoring external factors. We craft this probe for an agent $i$ by first extracting its assigned lane sequence in $\mathbf{X}^{(i)}_{hist}$. Next, we convert its coordinates to a Frenet frame~\cite{werling2010optimal}, a coordinate system representing progress and displacements along the given lanes' centerlines. Finally, we perform a constant-velocity extrapolation in the Frenet frame, to compute a ``future extrapolated'' trajectory. For agents without a lane assignment, we perform the same steps in Cartesian space. We denote this future extrapolated trajectory as $\mathbf{X}^{(i)}_{FE}$, in contrast with the original ground truth trajectory, $\mathbf{X}^{(i)}_{GT}$. 


To incorporate this method into the trajectory score in \Cref{eq:traj_score}, we extract the individual and social features of both $\mathbf{X}^{(i)}_{GT}$ and $\mathbf{X}^{(i)}_{FE}$. 
We first compute the individual score using $\mathbf{X}^{(i)}_{FE}$. 
To compute the social interaction scores, for a pair of interacting agents $(i, j)$, we compute $i$'s social score between $(\mathbf{X}^{(i)}_{FE}, \mathbf{X}^{(j)}_{GT})$ and $j$'s score analogously. We denote this \textit{asymmetric} score as \texttt{TrajScore}$^{(i)}_{AS}$. Similarly, we compute the reference ground truth score using exclusively the \textit{GT} trajectories for both agents and denote this as \texttt{TrajScore}$^{(i)}_{GT}$. 
We then take the maximum value of these two scores into a final \textit{asymmetric combined} trajectory score, \texttt{TrajScore}$^{(i)}_{AC}$. In \Cref{tab:score_variations}, we summarize these scoring variations and ablations.


We compute a \texttt{SceneScore} for these trajectory variations by utilizing the corresponding \texttt{TrajScore} (\exempli \texttt{SceneScore}$_{FE}$ uses \texttt{TrajScore}$_{FE}$ exclusively, etc.). As shown in \Cref{fig:score_dist}, this overall scene scoring method follows a long-tailed distribution as desired. The scores that incorporate future extrapolation have a much wider spread than just the ground truth, indicating a greater variety of scenarios captured.


\section{Downstream Tasks} \label{sec:downstream_tasks}


We showcase the utility of our scenario scores from \Cref{sec:score_scheme} by applying them to two downstream tasks: 1) creating a safety-informed distribution shift to better evaluate trajectory prediction models; and 2) leveraging the scores to conduct remediation on such models, reducing the incurred drop in performance. 

\subsection{Distribution Shift Creation}
\label{ssec:distribution_shift_creation}

We wish to evaluate and improve the robustness of trajectory prediction models when facing scenes more challenging/safety-critical than those on which they were trained. That is, we must split $\mathcal{S}$ in such a way that $\mathcal{S}_{ID}$ contains relatively low safety-criticality while $\mathcal{S}_{OOD}$ contains the most criticality. Thus, we propose the following approach of splitting $\mathcal{S}$ into the desired safety-informed subsets. 

First, we implement a simple uniform, random training/validation/test split to analyze behavior absent of a domain shift context: \texttt{Uniform}. Next, as a baseline, we implement the cluster-based domain identification schema from \cite{ye2023improving}, representing a recent approach for domain shift creation that focuses on other aspects of the scenarios instead of safety-relevance: \texttt{Clusters}.
Finally, we incorporate a safety-informed approach leveraging our schema described in \Cref{sec:score_scheme}: \texttt{Scoring}. We hold out the top 20\% scoring scenes as the test set, then randomly partition the remaining scenes into training and validation.

\subsection{Robust Trajectory Prediction}\label{ssec:remediation_approach}

We propose a remediation strategy leveraging the proposed scores in \Cref{sec:score_scheme} to increase downstream prediction model performance on challenging, more safety-relevant scenarios. Inspired by the difficulty-weighting of samples, as discussed in \cite{zhou2022understanding}, we utilize \texttt{TrajScore}$^{(i)}_{AC}$ for each agent $i$ to linearly weigh its contribution to a prediction model's loss function, out of the $N$ total agents in a mini-batch:
\begin{equation} \label{eq:loss_weight}
    \texttt{WeightedLoss}^{(i)} = \frac{1}{N}\sum_{i}^{N} \texttt{Loss}^{(i)} * \texttt{Score}^{(i)}_{AC}
\end{equation}

\Cref{eq:loss_weight} is then applied after computing the loss function for a given model, but before invoking the optimization pass. This encourages the model to not treat all scenarios and agents' trajectories as equal and to care about more safety-relevant situations.  Next, because the future-extrapolated score depends only on information available in $s_{hist}$, we can incorporate \texttt{TrajScore}$^{(i)}_{FE}$ into a model directly, to add a sense of counterfactual understanding to its inductive biases. We encode this score for each agent $i$ with a simple multilayer perceptron. Then, we concatenate this feature directly with the context encoding representation used in each model (\idest a function of trajectory histories, lane embedding, etc.) before passing it to the model's trajectory decoding stage.

We also propose to incorporate a collision-aware loss objective within each model. Many models in AD trajectory forecasting produce multi-modal futures, where they output $K$ possible future modes for each agent, along with a scalar, confidence value for each~\cite{shi2022motion, konev2022motioncnn, tang2022golfer}. We add in a cross-entropy (CE) loss objective upon these confidence values, where the ``correct'' mode is the mode that minimizes collisions with other agents' ground truth futures. 
In the case where a model already has a CE loss objective (\exempli to minimize the distance to the agent's ground truth future), we linearly weigh the two target values according to a regularization parameter.


\section{Experimental Setup} \label{sec:experimental_setup}



\label{sec:dataset_implementation}

\noindent\textbf{Dataset}:
We utilize WOMD~\cite{ettinger2021large} as an exemplar dataset to validate our approach, as it contains a particularly wide variety of scenarios. This variety is highlighted in terms of both geographic and roadway diversity, as well as scene complexity and traffic density~\cite{wang2024survey, liu2024survey}. 
We utilize a subset from the publicly available training and validation sets from WOMD, consisting of roughly $170k$ scenarios. We consider our three different data splits (\Cref{ssec:distribution_shift_creation})---\texttt{Uniform}, \texttt{Clusters}, \texttt{Scoring}---to create $\mathcal{S}_{ID}$ for training and validation (roughly $135k$ scenes), and $\mathcal{S}_{OOD}$ for testing (roughly $35k$ scenes).




\vspace{0.1cm}
\noindent\textbf{Baselines}:
We implement two representative baseline models to validate the efficacy of our distribution shift and remediation strategies. First, we include MTR~\cite{shi2022motion}, which, as of this writing, is the current top-performing model on WOMD leaderboards. 
Second, we implement a version of A-VRNN~\cite{monti2020dagnet}, where we utilize social pooling~\cite{alahi2016sociallstm} instead of a graph attention layer for the hidden state refinement. While both models are designed to be ``socially-aware,'' neither is explicitly structured to predict safe futures. 
We follow the same training procedure performed by MTR, where the models are trained for 30 epochs, and learning rate reduction begins after epoch 20.

As a baseline remediation strategy, we implement the Frenet-based domain normalization approach in \cite{ye2023improving}. This approach converts all coordinates into a trajectory's Frenet frame, before passing the coordinates to a trajectory prediction model. In order to obtain reasonable performance, we use both the Cartesian and Frenet coordinates \textit{together} via concatenation, rather than replacing the former with the latter. We then implement our proposed remediation approach, described in \Cref{ssec:remediation_approach} for both models.

\vspace{0.1cm}
\noindent\textbf{Metrics}: To measure safety-criticality, we use collision rate (CR), as the average number of collisions of each predicted trajectory to the ground truth of the other agents, as in \cite{kothari2021human}, where collisions with the same external agent over multiple timesteps only count once.
We also utilize standard trajectory prediction metrics, as used in the WOMD challenge, including Average Displacement Error (ADE) and Final Displacement Error (FDE). These two metrics are used in a best-of-$K$ manner to report the mode with the smallest distance to the ground-truth future, over all predicted timesteps, and just the final predicted timestep, respectively. Another important metric used is mean Average Precision (mAP). This metric categorizes predicted modes into buckets (\exempli straight, stationary, u-turn, etc.), and punishes mode collapse for overlapping predictions.

\begin{figure*}[t]
    \vspace{0.2cm}
    \centering
    \includegraphics[width=1.00\textwidth,trim={0cm 0cm -0cm -0cm},clip]{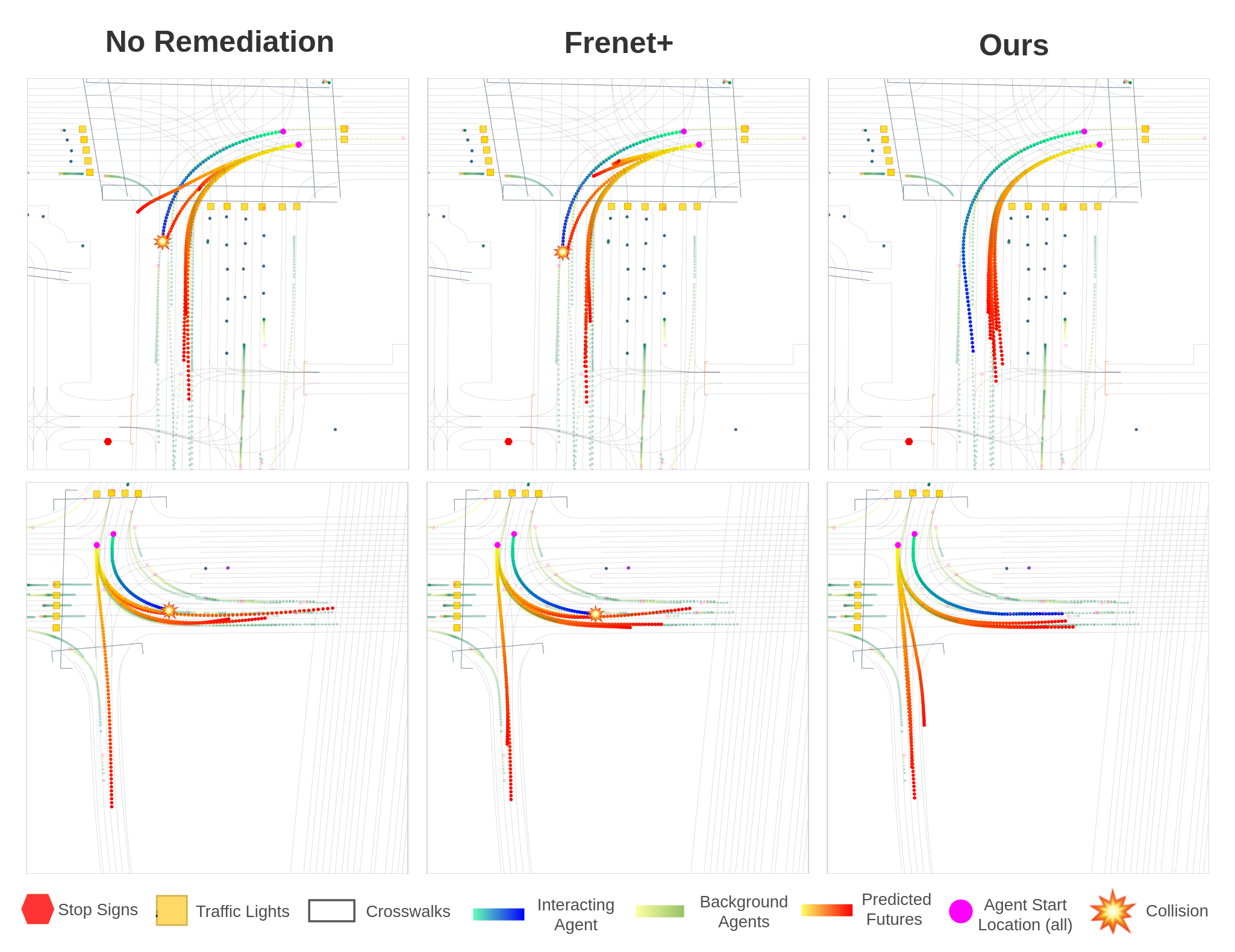}
    \caption{Qualitative examples of remediation approaches applied to MTR across two distinct scenarios. Trajectories progress from the pink starting points.}
    \vspace{-0.5cm}
    \label{fig:remediation_qualitative}
\end{figure*}

\section{Results} \label{sec:results}




\subsection{Distribution Shift Results}

In \Cref{fig:scored_scenes}, we highlight some examples of scenarios identified in $\mathcal{S}_{ID}$ and $\mathcal{S}_{OOD}$ for our \texttt{Scoring} method described in \Cref{sec:score_scheme}. The ID scenes contain both simple scenes with few interactions, as well as moderately safety-relevant scenes with lane changes and intersections. The OOD scenes appear significantly more safety-relevant, with more diverse maneuvers, such as u-turns, larger and more dangerous intersections, and many more VRUs navigating alongside vehicles. 

Our quantitative results for the trajectory prediction experiments are summarized in \Cref{tab:results_main_dist_shift}. The metric values reported are averaged over the three classes of vehicles, pedestrians, and cyclists. The $\Delta_{val}$ value in the final column indicates the increase in collision rate in the OOD test value compared to the ID validation value. In the \texttt{Uniform} split, as expected, results between $\mathcal{S}_{ID}$ and $\mathcal{S}_{OOD}$ are quite similar. For the \texttt{Clusters}~\cite{ye2023improving} split, we note that while a slight drop in metric performance for ADE / FDE and mAP occurred, the collision rate actually \textit{decreased} from validation to test. We suspect this is because the domains identified by this strategy have no sense of safety-criticality, affirming the importance of using such metrics when selecting scenes.
Finally, our \texttt{Scoring} strategy resulted in the largest increase in collision rates between $\mathcal{S}_{ID}$ and $\mathcal{S}_{OOD}$, both in terms of absolute value and percentage change. This increase occurs in both the ground truth tracks, as well as in our tested methods, more than doubling the in-distribution rates.


\begin{table*}[!ht]
\vspace{0.4cm}
\centering
\caption{Distribution shift experiments in WOMD~\cite{ettinger2021large}. ADE / FDE is in meters. $\Delta_{val}$ is the change in test collision rate (CR) from the corresponding val CR. A more drastic \textbf{increase} is better.} \label{tab:results_main_dist_shift}
\resizebox{\textwidth}{!}{%
\begin{tabular}{clcccccl}
\toprule
\multirow{2}{*}{\textbf{Data Split}} & \multirow{2}{*}{\textbf{Method}} & \multicolumn{3}{c}{\textbf{Validation Set (In-Distribution)}} & \multicolumn{3}{c}{\textbf{Testing Set (Out-of-Distribution)}}  \\ 
 & & ADE / FDE  & mAP  & CR  & ADE / FDE  & mAP  & \ \ CR \ ($\Delta_{val}$) \\ 
\midrule
\multirow{3}{*}{\texttt{Uniform}} 
     & GT  & - / - & - & 0.008 & - / - & - & 0.009 ($+12.5\%$) \\
     
     & MTR~\cite{shi2022motion}   & 0.73 / 1.58 & 0.30 & 0.062 & 0.73 / 1.59 & 0.31 & 0.061 ($-1.60\%$)\\
     & A-VRNN~\cite{monti2020dagnet}   & 1.80 / 4.63 & 0.06 & 0.057 & 1.82 / 4.67  & 0.06  & 0.058 ($+1.80\%$) \\
\midrule[0.7pt]

\multirow{3}{*}{\texttt{Clusters}~\cite{ye2023improving}} 
     & GT  & - / - & - & 0.009 & - / - & - & 0.007 ($-22.2\%$) \\
     
     & MTR & 0.69 / 1.50 & 0.35 & 0.060 & 0.71 / 1.55 & 0.33 & 0.051 ($-15.0\%$)\\
     & A-VRNN   & 1.79 / 4.59 & 0.08 & 0.062 & 1.82 / 4.70  & 0.07  & 0.049 ($-21.0\%$)\\
\midrule[0.7pt]


\multirow{3}{*}{\texttt{Scoring} (Ours)} 
     & GT  & - / - & - & 0.005 & - / - & - & \textbf{0.017} ($\mathbf{+240}\%$) \\
& MTR   &    0.72 / 1.59 & 0.32 & 0.044 & 0.74 / 1.59  & 0.30  & \textbf{0.100} ($\mathbf{+127}\%$) \\
& A-VRNN   &    1.99 / 5.26 & 0.05 & 0.042 & 2.13 / 5.55  & 0.05 & \textbf{0.099} ($\mathbf{+136}\%$) \\
\bottomrule
\end{tabular}
}
\begin{tablenotes}
\scriptsize
\item \textbf{GT}: Ground truth tracks
\end{tablenotes}
\vspace{-0.4cm}
\end{table*}

\begin{table*}[!h]
\vspace{0.6cm}
\centering
\caption{Robust trajectory prediction experiments in WOMD~\cite{ettinger2021large}. ADE / FDE is in meters. $\Delta_{test}$ is the change in test CR from the \textit{un-remediated} test CR for each method. A more drastic \textbf{decrease} is better.} \label{tab:results_main_remediation}
\resizebox{\textwidth}{!}{%
\begin{tabular}{clcccccl}
\toprule
\multirow{2}{*}{\textbf{Data Split}} & \multirow{2}{*}{\textbf{Method}} & \multicolumn{3}{c}{\textbf{Validation Set (In-Distribution)}} & \multicolumn{3}{c}{\textbf{Testing Set (Out-of-Distribution)}}  \\ 
 & & ADE / FDE  & mAP  & CR  & ADE / FDE  & mAP  & \ \ CR \ ($\Delta_{test}$) \\ 
\midrule

\multirow{7}{*}{\texttt{Scoring} (Ours)} 
     & GT  & - / - & - & 0.005 & - / - & - & 0.017 (\ \  \ \ \ \ - \ \ ) \\
 \cmidrule(l){2-8}
& MTR   &    0.72 / 1.59 & 0.32 & 0.044 & 0.74 / 1.59  & 0.30  & 0.100 (\ \  \ \ \ \ - \ \ ) \\
     & MTR + F+~\cite{ye2023improving} &       0.73 / 1.59 & 0.32 & 0.043 & 0.75 / 1.59  & 0.30  & 0.099 ($-1.00\%$)  \\
     & MTR + Ours &   0.83 / 1.80 & 0.25 & 0.037 & 0.89 / 1.91  & 0.22  & \textbf{0.086} ($\mathbf{-14.0\%}$) \\ 
     \cmidrule(l){2-8}
& A-VRNN   &    1.99 / 5.26 & 0.05 & 0.042 & 2.13 / 5.55  & 0.05 & 0.099 (\ \  \ \ \ \ - \ \ ) \\
     & A-VRNN + F+ &             2.05 / 5.24 & 0.06 & 0.041 & 2.23 / 5.73  & 0.06  & 0.103 ($+4.04\%$) \\
     &  A-VRNN + Ours &          1.76 / 4.61 & 0.06 & 0.039 & 1.91 / 4.94 & 0.06 & \textbf{0.093} ($\mathbf{-6.06\%}$) \\ 
\bottomrule
\end{tabular}
}
\begin{tablenotes}
\scriptsize
\item \textbf{GT}: Ground truth tracks, \textbf{F+}: Frenet+ Strategy~\cite{ye2023improving}
\end{tablenotes}
\vspace{-0.5cm}
\end{table*}

\subsection{Robust Trajectory Prediction Results}

We show our remediation experiment results in  \Cref{tab:results_main_remediation}. Our proposed method was the most effective in reducing collisions for the tested models, as shown by the $\Delta_{test}$ values. For MTR in particular, we observe the test collision rates are lowered by $14\%$, while for A-VRNN, the rates decrease by $6\%$. This resulted in an average decrease of $10\%$, reducing the gap to the ground truth collision rate. However, our method does result in a slight decrease in performance on other metrics for MTR. This is likely because MTR has an existing CE loss to select the best mode based on these other metrics, meaning the collision loss objective is in contention with its original objective. 

Furthermore, the Frenet+ strategy~\cite{ye2023improving} appeared ineffective in remediating the drop in performance on the \texttt{Scoring} data split. We suspect this is due to the presence of more object types than just vehicles; cyclists and pedestrians are often not in lanes, so incorporating such lane information may have been more harmful than beneficial. Additionally, even for vehicles following well-defined lanes, the Frenet+ strategy can still incur collisions, particularly at intersections and unprotected turns. 

To gain further insight into the benefits of both the Frenet+ strategy and our remediation approach, we provide qualitative examples in \Cref{fig:remediation_qualitative} using MTR as the prediction model. In these scenarios, the prediction with no remediation results in future modes that collide with an external agent. Meanwhile, the Frenet+ strategy is able to better stay in lanes than the un-remediated approach but still results in collisions. Finally, our remediation approach is able to avoid collisions, while still providing reasonable mode diversity and lane conformance.






\subsection{Ablation Studies}

As shown in \Cref{tab:results_scoring}, we performed a distribution shift ablation study focusing on the five variations of our scoring strategy discussed in \Cref{sec:score_scheme}. We utilized MTR as it is the best model according to traditional metrics. Our full method, with asymmetric combined scoring, resulted in the largest increase in collision rate, while still incurring a moderate increase in the other metrics. This result confirms our hypothesis from \Cref{ssec:counterfactual_rescoring} that our counterfactual probing technique indeed captures a fuller spectrum of safety-relevant scenes.

We also performed an ablation study focusing on aspects of our remediation strategy, as shown in \Cref{tab:results_remediation}. While the collision loss objective alone was quite effective, the best performance was achieved utilizing our full approach, incorporating the scores as part of the models' inductive biases and loss weights as well.

\begin{table*}[!ht]
\vspace{0.2cm}
\centering
\caption{Scoring strategy ablation study. Results are from using MTR~\cite{shi2022motion} on WOMD~\cite{ettinger2021large}. ADE / FDE is in meters. $\Delta_{val}$ is the change in test CR from val for the given method. The best distribution shift result is \textbf{bolded}.} \label{tab:results_scoring}
\resizebox{\textwidth}{!}{%
\begin{tabular}{lccccccccc}
\toprule
\multirow{2}{*}{\textbf{Ablation Name}} & \multicolumn{3}{c}{\textbf{Scoring Strategy}} & \multicolumn{3}{c}{\textbf{Validation Set (In-Distribution)}} & \multicolumn{3}{c}{\textbf{Testing Set (Out-of-Distribution)}}  \\ 
& GT & FE & AS & ADE / FDE  & mAP  & CR  & ADE / FDE  & mAP  & CR ($\Delta_{val}$)  \\ 
\midrule
    Ground Truth & $\checkmark$ & - & - & 0.72 / 1.57 & 0.32 & 0.041 & 0.75 / 1.64 & 0.29 & 0.088 ($+115\%$) \\
    Future Extrapolated & - & $\checkmark$ &  - & 0.73 / 1.61 & 0.33 & 0.046 & 0.72 / 1.55 & 0.31 & 0.097 ($+111\%$)\\
    Combined & $\checkmark$ & $\checkmark$ & - & 0.73 / 1.60 & 0.32 & 0.048 & 0.74 / 1.59 & 0.29 & 0.098 ($+104\%)$\\
    Asymmetric & - & $\checkmark$ & $\checkmark$ & 0.73 / 1.61 & 0.33 & 0.044 & 0.73 / 1.58 & 0.30 & 0.099 ($+125\%$)\\
    Asymmetric Combined & $\checkmark$ & $\checkmark$ & $\checkmark$ & 0.72 / 1.59 & 0.32 & 0.044 & 0.74 / 1.59  & 0.30  & \textbf{0.100 ($\mathbf{+127\%}$)} \\
\bottomrule
\end{tabular}
}
\begin{tablenotes}
\scriptsize
\item \textbf{GT}: Ground truth, \textbf{FE}: Future extrapolated, \textbf{AS}: Asymmetric scoring.
\end{tablenotes}
\vspace{-0.4cm}
\end{table*}

\begin{table*}[!ht]
\vspace{0.4cm}
\centering
\caption{Remediation strategy ablation study based on our proposed approach in \Cref{ssec:remediation_approach} utilizing MTR~\cite{shi2022motion} on WOMD~\cite{ettinger2021large}. ADE / FDE is in meters. $\Delta_{test}$ is the change in test CR from the \textit{un-remediated} MTR test CR.} \label{tab:results_remediation}
\resizebox{\textwidth}{!}{%
\begin{tabular}{lcccccccc@{}}
\toprule

\multirow{2}{*}{\textbf{Ablation Name}} & \multicolumn{2}{c}{\textbf{Remediation}} & \multicolumn{3}{c}{\textbf{Validation Set (In-Distribution)}} & \multicolumn{3}{c}{\textbf{Testing Set (Out-of-Distribution)}}  \\ 

  & SC & CL & ADE / FDE & mAP  & CR & ADE / FDE  & mAP & CR ($\Delta_{test}$) \\
  \midrule
   MTR~\cite{shi2022motion} & - & - & 0.72 / 1.59 & 0.32 & 0.044 & 0.74 / 1.59  & 0.30  & 0.100 (\ \ \ \ \ - \ \ \ ) \\
   MTR + Ours (SC only)  & $\checkmark$ & - & 0.74 / 1.63 & 0.31 & 0.046 & 0.74 / 1.61 & 0.29 & 0.103 ($+3.00\%$) \\
   MTR + Ours (CL only) & - & $\checkmark$ & 0.81 / 1.77 & 0.27 & 0.038 & 0.88 / 1.92 & 0.23 & 0.093 ($-7.00\%$) \\
   MTR + Ours (Full) & $\checkmark$ & $\checkmark$ & 0.83 / 1.80 & 0.25 & 0.037 & 0.89 / 1.91  & 0.22  & \textbf{0.086 ($\mathbf{-14.0}\%$)} \\
\bottomrule
\end{tabular}
}
\begin{tablenotes}
\scriptsize
\item \textbf{SC}: Score incorporation, \textbf{CL}: Collision loss objective.
\end{tablenotes}
\vspace{-0.5cm}
\end{table*}

\section{Conclusion} \label{sec:discussion}


Developing autonomous driving trajectory prediction models through real-world datasets, such as WOMD, is often considered insufficient for ensuring robustness and safety. While such datasets provide realistic recorded scenarios, they rarely contain truly safety-relevant scenarios, falling victim to the ``curse-of-rarity.'' Still, we proposed to further characterize these datasets and find hidden safety-relevant scenarios therein. 
We thus provided a versatile scenario characterization approach to score scenarios by a hierarchical combination of complementary individual and social features. By performing a counterfactual probe, emulating how a distracted agent may operate, we extended the spectrum of safety-relevance to additionally find hidden risky scenarios, without requiring unrealistic simulation or dangerous real-world testing. 

Under a distribution shift setting where the most safety-relevant scenes were held out as out-of-distribution, we demonstrated that both ground truth, as well as our evaluated trajectory prediction models, incurred a significant increase in collision rates. We further contributed a remediation strategy, achieving a 10\% average reduction in prediction collision rates.



Although our remediation strategy proved successful in reducing the test collision rate, the drop in performance was not remediated completely. Incorporating test-time refinement and collaborative sampling techniques, as highlighted in contemporaneous work, could prove a fruitful direction in improving this strategy further~\cite{kothari2023safety}. 
Another interesting future direction of this work would be to utilize our scoring strategy to assess safety-critical scenarios generated in simulation along the axes of realism, frequency, and type of safety-relevance created. Overall, we argue that trajectory prediction datasets can still be utilized in assessing safety in autonomous driving, and encourage future work to further this direction.

\balance


\bibliographystyle{IEEEtranBST/IEEEtran}
\bibliography{IEEEtranBST/IEEEabrv, ref}

\begin{thebibliography}{10}
\providecommand{\url}[1]{#1}
\csname url@rmstyle\endcsname
\providecommand{\newblock}{\relax}
\providecommand{\bibinfo}[2]{#2}
\providecommand\BIBentrySTDinterwordspacing{\spaceskip=0pt\relax}
\providecommand\BIBentryALTinterwordstretchfactor{4}
\providecommand\BIBentryALTinterwordspacing{\spaceskip=\fontdimen2\font plus
\BIBentryALTinterwordstretchfactor\fontdimen3\font minus \fontdimen4\font\relax}
\providecommand\BIBforeignlanguage[2]{{%
\expandafter\ifx\csname l@#1\endcsname\relax
\typeout{** WARNING: IEEEtran.bst: No hyphenation pattern has been}%
\typeout{** loaded for the language `#1'. Using the pattern for}%
\typeout{** the default language instead.}%
\else
\language=\csname l@#1\endcsname
\fi
#2}}

\bibitem{guo2022maturity}
X.~Guo and Y.~Zhang, ``Maturity in automated driving on public roads: a review of the six-year autonomous vehicle tester program,'' \emph{Transportation research record}, vol. 2676, no.~11, pp. 352--362, 2022.

\bibitem{francis2022learn}
J.~Francis, B.~Chen, S.~Ganju, S.~Kathpal, J.~Poonganam, A.~Shivani, V.~Vyas, S.~Genc, I.~Zhukov, M.~Kumskoy, \emph{et~al.}, ``Learn-to-race challenge 2022: Benchmarking safe learning and cross-domain generalisation in autonomous racing,'' \emph{arXiv preprint arXiv:2205.02953}, 2022.

\bibitem{teng2023motion}
S.~Teng, X.~Hu, P.~Deng, B.~Li, Y.~Li, Y.~Ai, D.~Yang, L.~Li, Z.~Xuanyuan, F.~Zhu, \emph{et~al.}, ``Motion planning for autonomous driving: The state of the art and future perspectives,'' \emph{IEEE Transactions on Intelligent Vehicles}, 2023.

\bibitem{kothari2023safety}
P.~Kothari and A.~Alahi, ``Safety-compliant generative adversarial networks for human trajectory forecasting,'' \emph{IEEE Transactions on Intelligent Transportation Systems}, vol.~24, no.~4, pp. 4251--4261, 2023.

\bibitem{ettinger2021large}
S.~Ettinger, S.~Cheng, B.~Caine, C.~Liu, H.~Zhao, S.~Pradhan, Y.~Chai, B.~Sapp, C.~R. Qi, Y.~Zhou, \emph{et~al.}, ``Large scale interactive motion forecasting for autonomous driving: The waymo open motion dataset,'' in \emph{Proceedings of the IEEE/CVF International Conference on Computer Vision}, 2021, pp. 9710--9719.

\bibitem{ding2020learning}
W.~Ding, B.~Chen, M.~Xu, and D.~Zhao, ``Learning to collide: An adaptive safety-critical scenarios generating method,'' in \emph{2020 IEEE/RSJ International Conference on Intelligent Robots and Systems (IROS)}.\hskip 1em plus 0.5em minus 0.4em\relax IEEE, 2020, pp. 2243--2250.

\bibitem{ding2023survey}
W.~Ding, C.~Xu, M.~Arief, H.~Lin, B.~Li, and D.~Zhao, ``A survey on safety-critical driving scenario generation—a methodological perspective,'' \emph{IEEE Transactions on Intelligent Transportation Systems}, 2023.

\bibitem{feng2023dense}
S.~Feng, H.~Sun, X.~Yan, H.~Zhu, Z.~Zou, S.~Shen, and H.~X. Liu, ``Dense reinforcement learning for safety validation of autonomous vehicles,'' \emph{Nature}, vol. 615, no. 7953, pp. 620--627, 2023.

\bibitem{webb2020waymo}
N.~Webb, D.~Smith, C.~Ludwick, T.~Victor, Q.~Hommes, F.~Favaro, G.~Ivanov, and T.~Daniel, ``Waymo's safety methodologies and safety readiness determinations,'' \emph{arXiv preprint arXiv:2011.00054}, 2020.

\bibitem{huang2016autonomous}
W.~Huang, K.~Wang, Y.~Lv, and F.~Zhu, ``Autonomous vehicles testing methods review,'' in \emph{2016 IEEE 19th International Conference on Intelligent Transportation Systems (ITSC)}.\hskip 1em plus 0.5em minus 0.4em\relax IEEE, 2016, pp. 163--168.

\bibitem{francis2022core}
J.~Francis, N.~Kitamura, F.~Labelle, X.~Lu, I.~Navarro, and J.~Oh, ``Core challenges in embodied vision-language planning,'' \emph{Journal of Artificial Intelligence Research}, vol.~74, pp. 459--515, 2022.

\bibitem{huang2023went}
P.~Huang, X.~Zhang, Z.~Cao, S.~Liu, M.~Xu, W.~Ding, J.~Francis, B.~Chen, and D.~Zhao, ``What went wrong? closing the sim-to-real gap via differentiable causal discovery,'' in \emph{Conference on Robot Learning}.\hskip 1em plus 0.5em minus 0.4em\relax PMLR, 2023, pp. 734--760.

\bibitem{xu2023bits}
D.~Xu, Y.~Chen, B.~Ivanovic, and M.~Pavone, ``Bits: Bi-level imitation for traffic simulation,'' in \emph{2023 IEEE International Conference on Robotics and Automation (ICRA)}.\hskip 1em plus 0.5em minus 0.4em\relax IEEE, 2023, pp. 2929--2936.

\bibitem{cao2023robust}
Y.~Cao, D.~Xu, X.~Weng, Z.~Mao, A.~Anandkumar, C.~Xiao, and M.~Pavone, ``Robust trajectory prediction against adversarial attacks,'' in \emph{Conference on Robot Learning}.\hskip 1em plus 0.5em minus 0.4em\relax PMLR, 2023, pp. 128--137.

\bibitem{suo2023mixsim}
S.~Suo, K.~Wong, J.~Xu, J.~Tu, A.~Cui, S.~Casas, and R.~Urtasun, ``Mixsim: A hierarchical framework for mixed reality traffic simulation,'' in \emph{Proceedings of the IEEE/CVF Conference on Computer Vision and Pattern Recognition}, 2023, pp. 9622--9631.

\bibitem{shalev2017formal}
S.~Shalev-Shwartz, S.~Shammah, and A.~Shashua, ``On a formal model of safe and scalable self-driving cars,'' \emph{arXiv preprint arXiv:1708.06374}, 2017.

\bibitem{filos2020can}
A.~Filos, P.~Tigkas, R.~McAllister, N.~Rhinehart, S.~Levine, and Y.~Gal, ``Can autonomous vehicles identify, recover from, and adapt to distribution shifts?'' in \emph{International Conference on Machine Learning}.\hskip 1em plus 0.5em minus 0.4em\relax PMLR, 2020, pp. 3145--3153.

\bibitem{ye2023improving}
L.~Ye, Z.~Zhou, and J.~Wang, ``Improving the generalizability of trajectory prediction models with frenet-based domain normalization,'' \emph{arXiv preprint arXiv:2305.17965}, 2023.

\bibitem{itkina2023interpretable}
M.~Itkina and M.~Kochenderfer, ``Interpretable self-aware neural networks for robust trajectory prediction,'' in \emph{Conference on Robot Learning}.\hskip 1em plus 0.5em minus 0.4em\relax PMLR, 2023, pp. 606--617.

\bibitem{moers2022exid}
T.~Moers, L.~Vater, R.~Krajewski, J.~Bock, A.~Zlocki, and L.~Eckstein, ``The exid dataset: A real-world trajectory dataset of highly interactive highway scenarios in germany,'' in \emph{2022 IEEE Intelligent Vehicles Symposium (IV)}.\hskip 1em plus 0.5em minus 0.4em\relax IEEE, 2022, pp. 958--964.

\bibitem{glasmacher2022automated}
C.~Glasmacher, R.~Krajewski, and L.~Eckstein, ``An automated analysis framework for trajectory datasets,'' \emph{arXiv preprint arXiv:2202.07438}, 2022.

\bibitem{sadat2021diverse}
A.~Sadat, S.~Segal, S.~Casas, J.~Tu, B.~Yang, R.~Urtasun, and E.~Yumer, ``Diverse complexity measures for dataset curation in self-driving,'' in \emph{2021 IEEE/RSJ International Conference on Intelligent Robots and Systems (IROS)}.\hskip 1em plus 0.5em minus 0.4em\relax IEEE, 2021, pp. 8609--8616.

\bibitem{rudenko2020human}
A.~Rudenko, L.~Palmieri, M.~Herman, K.~M. Kitani, D.~M. Gavrila, and K.~O. Arras, ``Human motion trajectory prediction: A survey,'' \emph{The International Journal of Robotics Research}, vol.~39, no.~8, pp. 895--935, 2020.

\bibitem{monti2020dagnet}
A.~Monti, A.~Bertugli, S.~Calderara, and R.~Cucchiara, ``Dag-net: Double attentive graph neural network for trajectory forecasting,'' in \emph{25th International Conference on Pattern Recognition, {ICPR} 2020, Virtual Event / Milan, Italy, January 10-15, 2021}.\hskip 1em plus 0.5em minus 0.4em\relax {IEEE}, 2020, pp. 2551--2558.

\bibitem{sadeghian2019sophie}
A.~Sadeghian, V.~Kosaraju, A.~Sadeghian, N.~Hirose, H.~Rezatofighi, and S.~Savarese, ``Sophie: An attentive {GAN} for predicting paths compliant to social and physical constraints,'' in \emph{{IEEE} Conference on Computer Vision and Pattern Recognition, {CVPR} 2019, Long Beach, CA, USA, June 16-20, 2019}.\hskip 1em plus 0.5em minus 0.4em\relax Computer Vision Foundation / {IEEE}, 2019, pp. 1349--1358.

\bibitem{park2020diverse}
S.~H. Park, G.~Lee, J.~Seo, M.~Bhat, M.~Kang, J.~Francis, A.~Jadhav, P.~P. Liang, and L.-P. Morency, ``Diverse and admissible trajectory forecasting through multimodal context understanding,'' in \emph{Computer Vision--ECCV 2020: 16th European Conference, Glasgow, UK, August 23--28, 2020, Proceedings, Part XI 16}.\hskip 1em plus 0.5em minus 0.4em\relax Springer, 2020, pp. 282--298.

\bibitem{shi2022motion}
S.~Shi, L.~Jiang, D.~Dai, and B.~Schiele, ``Motion transformer with global intention localization and local movement refinement,'' \emph{Advances in Neural Information Processing Systems}, vol.~35, pp. 6531--6543, 2022.

\bibitem{bhat2020trajformer}
M.~Bhat, J.~Francis, and J.~Oh, ``Trajformer: Trajectory prediction with local self-attentive contexts for autonomous driving,'' \emph{arXiv preprint arXiv:2011.14910}, 2020.

\bibitem{tang2022golfer}
X.~Tang, S.~S. Eshkevari, H.~Chen, W.~Wu, W.~Qian, and X.~Wang, ``Golfer: Trajectory prediction with masked goal conditioning mnm network,'' \emph{arXiv preprint arXiv:2207.00738}, 2022.

\bibitem{alahi2016sociallstm}
A.~Alahi, K.~Goel, V.~Ramanathan, A.~Robicquet, L.~Fei{-}Fei, and S.~Savarese, ``Social {LSTM:} human trajectory prediction in crowded spaces,'' in \emph{2016 {IEEE} Conference on Computer Vision and Pattern Recognition, {CVPR} 2016, Las Vegas, NV, USA, June 27-30, 2016}.\hskip 1em plus 0.5em minus 0.4em\relax {IEEE} Computer Society, 2016, pp. 961--971.

\bibitem{konev2022motioncnn}
S.~Konev, K.~Brodt, and A.~Sanakoyeu, ``Motioncnn: A strong baseline for motion prediction in autonomous driving,'' 2022.

\bibitem{yan2021agentformer}
Y.~Yuan, X.~Weng, Y.~Ou, and K.~Kitani, ``Agentformer: Agent-aware transformers for socio-temporal multi-agent forecasting,'' \emph{CoRR}, vol. abs/2103.14023, 2021.

\bibitem{ngiam2021scene}
J.~Ngiam, V.~Vasudevan, B.~Caine, Z.~Zhang, H.-T.~L. Chiang, J.~Ling, R.~Roelofs, A.~Bewley, C.~Liu, A.~Venugopal, \emph{et~al.}, ``Scene transformer: A unified architecture for predicting future trajectories of multiple agents,'' in \emph{International Conference on Learning Representations}, 2021.

\bibitem{liang2020learning}
M.~Liang, B.~Yang, R.~Hu, Y.~Chen, R.~Liao, S.~Feng, and R.~Urtasun, ``Learning lane graph representations for motion forecasting,'' in \emph{Computer Vision--ECCV 2020: 16th European Conference, Glasgow, UK, August 23--28, 2020, Proceedings, Part II 16}.\hskip 1em plus 0.5em minus 0.4em\relax Springer, 2020, pp. 541--556.

\bibitem{gao2020vectornet}
J.~Gao, C.~Sun, H.~Zhao, Y.~Shen, D.~Anguelov, C.~Li, and C.~Schmid, ``Vectornet: Encoding hd maps and agent dynamics from vectorized representation,'' in \emph{Proceedings of the IEEE/CVF Conference on Computer Vision and Pattern Recognition}, 2020, pp. 11\,525--11\,533.

\bibitem{weng2023joint}
E.~Weng, H.~Hoshino, D.~Ramanan, and K.~Kitani, ``Joint metrics matter: {A} better standard for trajectory forecasting,'' \emph{CoRR}, vol. abs/2305.06292, 2023.

\bibitem{saadatnejad2022socially}
S.~Saadatnejad, M.~Bahari, P.~Khorsandi, M.~Saneian, S.-M. Moosavi-Dezfooli, and A.~Alahi, ``Are socially-aware trajectory prediction models really socially-aware?'' \emph{Transportation research part C: emerging technologies}, vol. 141, p. 103705, 2022.

\bibitem{zhang2022adversarial}
Q.~Zhang, S.~Hu, J.~Sun, Q.~A. Chen, and Z.~M. Mao, ``On adversarial robustness of trajectory prediction for autonomous vehicles,'' in \emph{Proceedings of the IEEE/CVF Conference on Computer Vision and Pattern Recognition}, 2022, pp. 15\,159--15\,168.

\bibitem{hanselmann2022king}
N.~Hanselmann, K.~Renz, K.~Chitta, A.~Bhattacharyya, and A.~Geiger, ``King: Generating safety-critical driving scenarios for robust imitation via kinematics gradients,'' in \emph{European Conference on Computer Vision}.\hskip 1em plus 0.5em minus 0.4em\relax Springer, 2022, pp. 335--352.

\bibitem{werling2010optimal}
M.~Werling, J.~Ziegler, S.~Kammel, and S.~Thrun, ``Optimal trajectory generation for dynamic street scenarios in a frenet frame,'' in \emph{2010 IEEE international conference on robotics and automation}.\hskip 1em plus 0.5em minus 0.4em\relax IEEE, 2010, pp. 987--993.

\bibitem{kothari2023motion}
P.~Kothari, D.~Li, Y.~Liu, and A.~Alahi, ``Motion style transfer: Modular low-rank adaptation for deep motion forecasting,'' in \emph{Conference on Robot Learning}.\hskip 1em plus 0.5em minus 0.4em\relax PMLR, 2023, pp. 774--784.

\bibitem{chang2019argoverse}
M.-F. Chang, J.~Lambert, P.~Sangkloy, J.~Singh, S.~Bak, A.~Hartnett, D.~Wang, P.~Carr, S.~Lucey, D.~Ramanan, \emph{et~al.}, ``Argoverse: 3d tracking and forecasting with rich maps,'' in \emph{Proceedings of the IEEE/CVF conference on computer vision and pattern recognition}, 2019, pp. 8748--8757.

\bibitem{caesar2020nuscenes}
H.~Caesar, V.~Bankiti, A.~H. Lang, S.~Vora, V.~E. Liong, Q.~Xu, A.~Krishnan, Y.~Pan, G.~Baldan, and O.~Beijbom, ``nuscenes: A multimodal dataset for autonomous driving,'' in \emph{Proceedings of the IEEE/CVF conference on computer vision and pattern recognition}, 2020, pp. 11\,621--11\,631.

\bibitem{zhang2021finding}
X.~Zhang, J.~Tao, K.~Tan, M.~T{\"o}rngren, J.~M.~G. S{\'a}nchez, M.~R. Ramli, X.~Tao, M.~Gyllenhammar, F.~Wotawa, N.~Mohan, \emph{et~al.}, ``Finding critical scenarios for automated driving systems: A systematic literature review,'' \emph{arXiv preprint arXiv:2110.08664}, 2021.

\bibitem{weber2019framework}
H.~Weber, J.~Bock, J.~Klimke, C.~Rösener, J.~Hiller, R.~Krajewski, A.~Zlocki, and L.~Eckstein, ``A framework for definition of logical scenarios for safety assurance of automated driving,'' \emph{Traffic Injury Prevention}, vol.~20, pp. S65--S70, 06 2019.

\bibitem{schmidt2022meat}
J.~Schmidt, J.~Jordan, D.~Raba, T.~Welz, and K.~Dietmayer, ``Meat: Maneuver extraction from agent trajectories,'' in \emph{2022 IEEE Intelligent Vehicles Symposium (IV)}.\hskip 1em plus 0.5em minus 0.4em\relax IEEE, 2022, pp. 1810--1816.

\bibitem{guha2022robust}
A.~Guha, R.~Lei, J.~Zhu, X.~Nguyen, and D.~Zhao, ``Robust unsupervised learning of temporal dynamic vehicle-to-vehicle interactions,'' \emph{Transportation research part C: emerging technologies}, vol. 142, p. 103768, 2022.

\bibitem{zhan2019interaction}
W.~Zhan, L.~Sun, D.~Wang, H.~Shi, A.~Clausse, M.~Naumann, J.~Kummerle, H.~Konigshof, C.~Stiller, A.~de~La~Fortelle, \emph{et~al.}, ``Interaction dataset: An international, adversarial and cooperative motion dataset in interactive driving scenarios with semantic maps,'' \emph{arXiv preprint arXiv:1910.03088}, 2019.

\bibitem{vogel2003comparison}
K.~Vogel, ``A comparison of headway and time to collision as safety indicators,'' \emph{Accident Analysis \& Prevention}, vol.~35, no.~3, pp. 427--433, 2003.

\bibitem{shen2020crash}
J.~Shen and G.~Yang, ``Crash risk assessment for heterogeneity traffic and different vehicle-following patterns using microscopic traffic flow data,'' \emph{Sustainability}, vol.~12, no.~23, p. 9888, 2020.

\bibitem{zhou2022understanding}
X.~Zhou, O.~Wu, W.~Zhu, and Z.~Liang, ``Understanding difficulty-based sample weighting with a universal difficulty measure,'' in \emph{Joint European Conference on Machine Learning and Knowledge Discovery in Databases}.\hskip 1em plus 0.5em minus 0.4em\relax Springer, 2022, pp. 68--84.

\bibitem{wang2024survey}
Y.~Wang, Z.~Han, Y.~Xing, S.~Xu, and J.~Wang, ``A survey on datasets for the decision making of autonomous vehicles,'' \emph{IEEE Intelligent Transportation Systems Magazine}, 2024.

\bibitem{liu2024survey}
M.~Liu, E.~Yurtsever, X.~Zhou, J.~Fossaert, Y.~Cui, B.~L. Zagar, and A.~C. Knoll, ``A survey on autonomous driving datasets: Data statistic, annotation, and outlook,'' \emph{arXiv preprint arXiv:2401.01454}, 2024.

\bibitem{kothari2021human}
P.~Kothari, S.~Kreiss, and A.~Alahi, ``Human trajectory forecasting in crowds: A deep learning perspective,'' \emph{IEEE Transactions on Intelligent Transportation Systems}, vol.~23, no.~7, pp. 7386--7400, 2021.

\end{thebibliography}
\end{document}